\documentclass[conference]{IEEEtran}
\IEEEoverridecommandlockouts
\usepackage{cite}
\usepackage{amsmath,amssymb,amsfonts}
\usepackage{algorithm}
\usepackage{graphicx}
\usepackage{textcomp}
\usepackage{xcolor}
\usepackage{mathtools}
\usepackage[noend]{algpseudocode}
\newcommand{\sig}{\sigma(k)}

\usepackage{amssymb,amsmath,amsthm}

\newtheorem{dfn}{Definition}
\usepackage{thmtools,thm-restate}
\usepackage{float}
\usepackage{subfig}
\def\BibTeX{{\rm B\kern-.05em{\sc i\kern-.025em b}\kern-.08em
    T\kern-.1667em\lower.7ex\hbox{E}\kern-.125emX}}

\begin{document}

\title{K-means for Evolving Data Streams}

\author{\IEEEauthorblockN{1\textsuperscript{st} Arkaitz Bidaurrazaga}
\IEEEauthorblockA{\textit{Basque Center for Applied Mathematics} \\
Bilbao, Spain \\
abidaurrazaga@bcamath.org}
\and
\IEEEauthorblockN{2\textsuperscript{nd} Aritz Pérez}
\IEEEauthorblockA{\textit{Basque Center for Applied Mathematics} \\
Bilbao, Spain \\
aperez@bcamath.org}
\and
\IEEEauthorblockN{3\textsuperscript{rd} Marco Capó}
\IEEEauthorblockA{\textit{Basque Center for Applied Mathematics} \\
Bilbao, Spain \\
mcapo@bcamath.org}}

\maketitle

\begin{abstract}
Nowadays, streaming data analysis has become a relevant area of research in machine learning. Most of the data streams available are unlabeled, and thus it is necessary to develop specific clustering techniques that take into account the particularities of the streaming data. In streaming data scenarios, the data is composed of an increasing sequence of batches of samples where the concept drift phenomenon may occur. In this work, we formally define the streaming $K$-means (S$K$M) problem, which implies a restart of the error function when a concept drift occurs. An approximated error function that does not rely on concept drift detection is proposed. We prove that such a surrogate is a good approximation of the S$K$M error. Then, we introduce an algorithm to deal with S$K$M problem by minimizing the surrogate error function each time a new batch arrives. Alternative initialization criteria are presented and theoretically analyzed for streaming data scenarios. Among them, we develop and analyze theoretically two initialization methods that search for the best trade-off between the importance that is given to the past and the current batches. The experiments show that the proposed algorithm with, the proposed initialization criteria, obtain the best results when dealing with the S$K$M problem without requiring to detect when concept drift takes place. 
\end{abstract}

\section{Introduction}

One of the most relevant unsupervised data analysis problems is clustering \cite{clusteringReview}, which consists of partitioning the data into a number of disjoint subsets called clusters. Among a wide variety of clustering methods, $K$-means algorithm is one of the most popular \cite{KmeansUsage}. In fact, it has been identified as one of the top-10 most important algorithms in data mining \cite{top10}. Before explaining the $K$-means algorithm, we shall briefly introduce the $K$-means problem. 

\subsection{$K$-means Problem}
Given a data set of $d$-dimensional points of size $n$, $X=\{\textbf{x}_i\}_{i=1}^{n} \subset \mathbb{R}^d$, the $K$-means problem is defined as finding a set of $K$ centroids $C=\{\textbf{c}_k\}_{k=1}^{K} \subset \mathbb{R}^d$, which minimizes the \textit{$K$-means error function}: 
\begin{equation}\label{eq:errf}
E(X,C)= \frac{1}{|X|}\cdot \sum_{x \in X} \|\textbf{x}-\textbf{c}_x\|^2\ ;\ \textbf{c}_x=\underset{\textbf{c}\in C}{\arg\min}\|\textbf{x}-\textbf{c}\|,
\end{equation}
where $\|\cdot\|$ denotes the Euclidean distance or $L^2$ norm.

One of the open questions in $K$-means is how to determine a proper value $K$ for each dataset. In the literature, one can find different techniques to approach this issue. In particular, \cite{pelleg2000x} propose a novel technique that determines a number of clusters by performing recursive bisections of a given set of clusters as long as the obtained Bayesian Information Criterion (BIC) is improved. On the other hand, \cite{hamerly2004learning} propose a method to learn $K$ based on a statistical test for determining whether instances are a random sample from a Gaussian distribution with arbitrary dimension and covariance matrix. Other approaches include  the use of cluster validity measures, such
as the Silhouette index, Davies-Bouldin and Calinski-Harabasz measures to automate the selection of this parameter \cite{hamalainen2017comparison,yuan2019research}.
In any case, in this work, as it is common in most of the $K$-means problem-related literature \cite{kmc2,SplitMerge,km++,makarychev2020improved}, the number of clusters is assumed to be predetermined.

\paragraph{$K$-means Algorithm} The $K$-means problem is known to be NP-hard for $K>1$ and $d>1$ \cite{NPhard}. The most popular heuristic approach to this problem is Lloyd's algorithm \cite{Lloyd}. Given a set of initial centroids, Lloyd's algorithm iterates two steps until convergence: 1) assignation step and 2) update step. In the assignation step, given a set of centroids, $C=\{\textbf{c}_k\}_{k=1}^K$, the set of points is partitioned into $K$ clusters, $\mathcal{P}=\{P_k\}_{k=1}^K$, by assigning each point to the closest centroid. Next, the new set of centroids is obtained by computing the center of mass of the points in each partition. This set of centroids minimizes the $K$-means error with respect to the given partition of the set of points. These two steps are repeated until a fixed point is reached, meaning, when the assignation step does not change the partition. This process has a $\mathcal{O}(n\cdot K \cdot d)$ time complexity. The combination of an \textit{initialization method }plus Lloyd's algorithm is called the $K$-means algorithm. Many alternative initialization methods exist and, in the upcoming section, we elaborate on them.

\paragraph{$K$-means Initialization} The solution obtained by $K$-means algorithm strongly depends on the initial set of centroids \cite{Init3,Init4,Init1}. Consequently, in the literature different initializations have been proposed. One of the most simple, yet effective, methods is Forgy's approach \cite{Forgy}. Forgy's initialization consists of choosing $K$ data points at random as initial centroids. The main drawback of this approach is that it tends to choose data points located at dense regions of the space, thus these regions tend to be over-represented. In order to address this problem, probabilistic methods with strong theoretical guarantees have been proposed. $K$-means++ (KM++) \cite{km++} initialization iteratively selects points from $X$ at random, where the probability of selection is proportional to the distance of the closest centroid previously selected. This strategy has become one of the most popular due its strong theoretical guarantees. Among other popular alternatives, we have variations of KM++, such as the $K$-means|| \cite{makarychev2020improved} and the Markov Chain KM++ \cite{kmc2}.

\subsection{Streaming Data}\label{sec:Related Work}

 Although the $K$-means problem deals with a fixed data set $X$, its usage can be generalized to scenarios in which data evolves over time. We define streaming data (SD) as a set of data batches that arrive sequentially, where each batch is a set of $d$-dimensional points. One of the main concerns when processing SD is how much data to store, since the volume of data increases indefinitely. Normally, a maximum number of stored batches is determined, this way time consumption and computational load of the clustering algorithm is controlled, and makes clustering tractable in this situation.

 Another main issue when dealing with SD is the \textit{concept drift} phenomenon. A concept drift occurs when the underlying probability distribution, associated to the batches, changes. In the presence of concept drifts we distinguish between passive and active approaches. On one hand, an active mechanism dynamically adjusts stored batches depending on whether a concept drift has occurred or not. On the other hand, in the passive approaches, more importance is given to recent batches. A detailed review on active and passive strategies to deal with concept drift can be found in \cite{gama2014survey}.
 
\subsection{Contributions}
In this paper, we formally define the Streaming $K$-means (S$K$M) problem. We describe an algorithm that does not need to detect concept drifts in order to obtain a good clustering. This approximation deals with the concept drift phenomenon by assigning exponentially decaying weights to older batches. We prove that the surrogate error is a good approximation to the S$K$M error, using Hoeffding's inequality \cite{Hoeffding}. The surrogate error is a weighted $K$-means error which gives higher weight to recent batches. Due to the importance of the initialization in the behaviour of weighted $K$-means, we present two initialization techniques that search for the best combination between past and novel information of clusters. We conduct experiments to compare them with other two straight-forward initialization strategies, which are to compute $K$M++ on the novel batch to obtain initial centroids or simply use previously computed centroids.

This paper is organized in the following way. In section \ref{SKM problem} the S$K$M problem is defined. Next, we propose a passive approach, and prove its suitability to deal with the S$K$M problem. In section \ref{sec:Initialization} we propose some appropriate initialization methods for the S$K$M problem. Finally, we conduct the experiments\footnote{Code available at
\textit{https://github.com/arkano29/Kmeans\_Streaming\_Data}} in section \ref{Experiments} to compare and discuss the results obtained for each method under different scenarios.  

\section{Streaming $K$-means Problem}\label{SKM problem}

In this section, we define the S$K$M problem, a natural adaptation of the $K$-means problem for SD consisting of the minimization of the S$K$M error. We define the S$K$M error function as follows:

\begin{dfn} \label{Priv}
Given a set of batches, $\mathcal{X}= \{B^t\}_{t\geq0}$ and a set of centroids $C$, the S$K$M error function is defined as
\begin{align}\label{eq:RealSKMError}
E_{*}(\mathcal{X},C)
=\frac{1}{M_T}\cdot \sum_{t=0}^{T-1}\sum_{x \in B^t} \|\textbf{x}-\textbf{c}_x\|^2,
\end{align}
where the index $t$ describes the antiquity of each batch, thus $B^0$ is the latest batch received, and $B^{T-1}$ represent the batch in which the last concept drift occurred, i.e., every batch $\{B^t\}_{t=0}^{T-1}$ shares the same underlying distribution. $M_T=\sum\limits_{t\leq T-1}|B^t|$ is the sum of each batch size.
\end{dfn}

In order to compute the S$K$M error function, we need to know the batch in which the last concept drift occurred,  $B^{T-1}$. The complexity of computing the S$K$M error is $\mathcal{O}(M_T\cdot K\cdot d)$. Clearly, the performance of an active approach to the problem will strongly depend on the behavior of the detection mechanism implemented. On one hand, when a fake drift is detected the past batches and the last centroids are discarded unnecessarily. On the other hand, if a concept drift occurs but is not detected, then the use of previous computed centroids could lead to a poor initialization and may infer a bad clustering. In this work, we describe an active algorithm, which we call Privileged S$K$M algorithm (PS$K$M). PS$K$M is an ideal baseline to the problem which knows in advance if a concept drift occurs (See Algorithm \ref{alg:IdealAlgorithma}), thus being able to compute and minimize the S$K$M error function. Alternatively, in the following section we propose a passive mechanism to approximate the solution of the S$K$M problem. PS$K$M will be the reference in the experimental section since we will simulate streaming data with concept drifts, and thus we will be able to compute the S$K$M error.

\begin{algorithm}[!t]
\caption{Privileged S$K$M algorithm (PS$K$M)}\label{alg:IdealAlgorithma}
\begin{algorithmic}[1]

\State \textbf{Input:} The set of previous batches $\mathcal{X}$, the obtained previous centroids $C$, and the new batch $B^0$.
\State \textbf{Output:} A set of centroids $C^*$, and its associated partition $\mathcal{P}^*$.

\If{$\mbox{concept drift}$}
\State $\mbox{Reset } \mathcal{X}$
\State $C \gets \mbox{$K$M++($B^0$)}$
\EndIf

\State $\mathcal{X} \gets \mbox{Append } B^0$
\State $C^*,\mathcal{P}^* \gets \mbox{Lloyd}(\mathcal{X},C)$
\State\Return $C^*,\mathcal{P}^*$

\end{algorithmic}
\end{algorithm}

\subsection{A Surrogate for S$K$M Error}\label{Surrogate function}

Here we propose a \textit{surrogate error function} for the S$K$M error function. This alternative function incorporates a forgetting mechanism based on a memory parameter, $\rho$, which assigns an exponentially decreasing weight $\rho^t$ to a given batch based on its antiquity $t$. Note that $t=0$ indicates the last batch obtained from the stream which has weight $1$. In particular, the approximated error function is defined as follows:

\begin{figure*}[!htbp]
	\centering
	\subfloat[]{\includegraphics[width=0.333\textwidth]{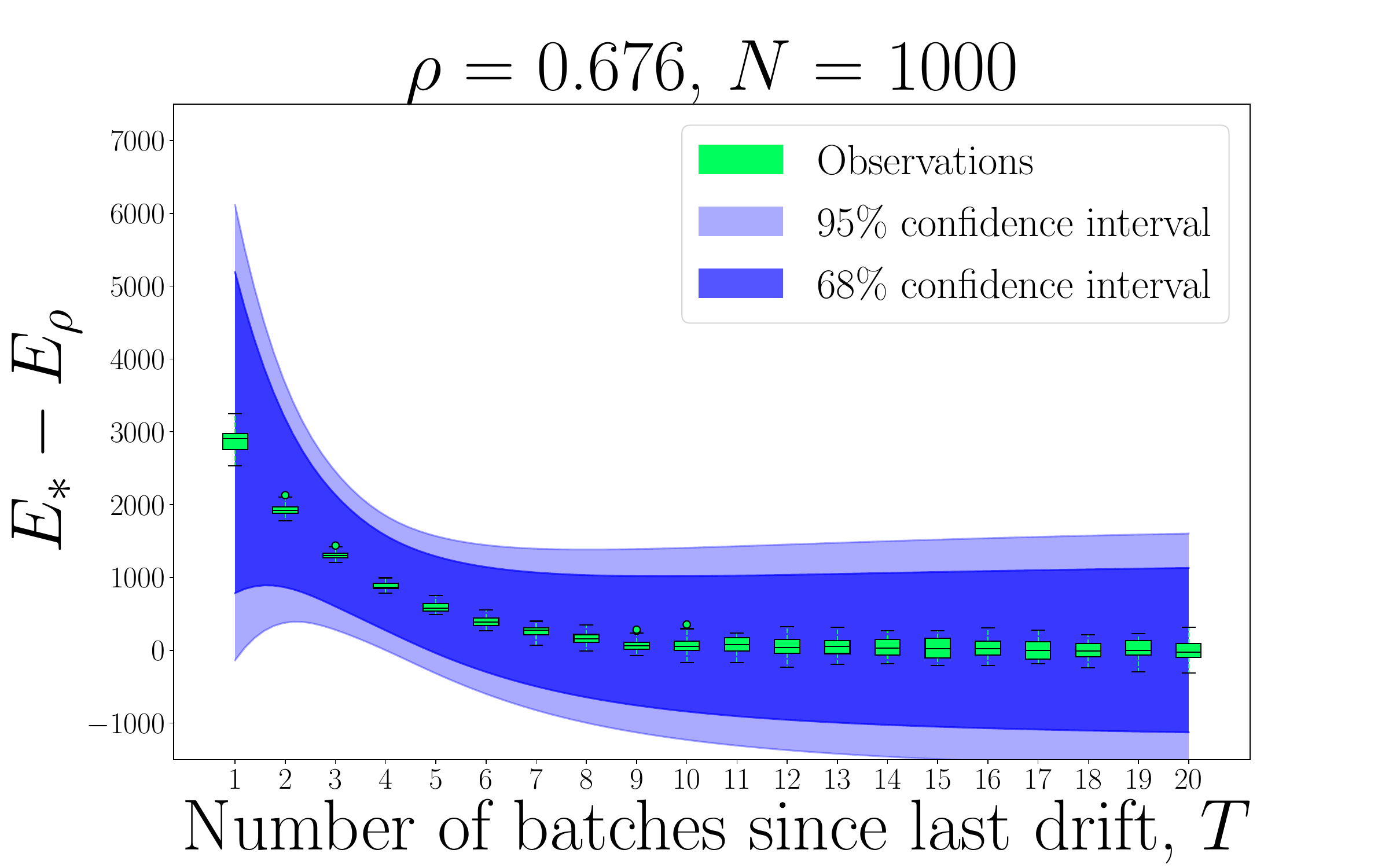}}
	\subfloat[]{\includegraphics[width=0.333\textwidth]{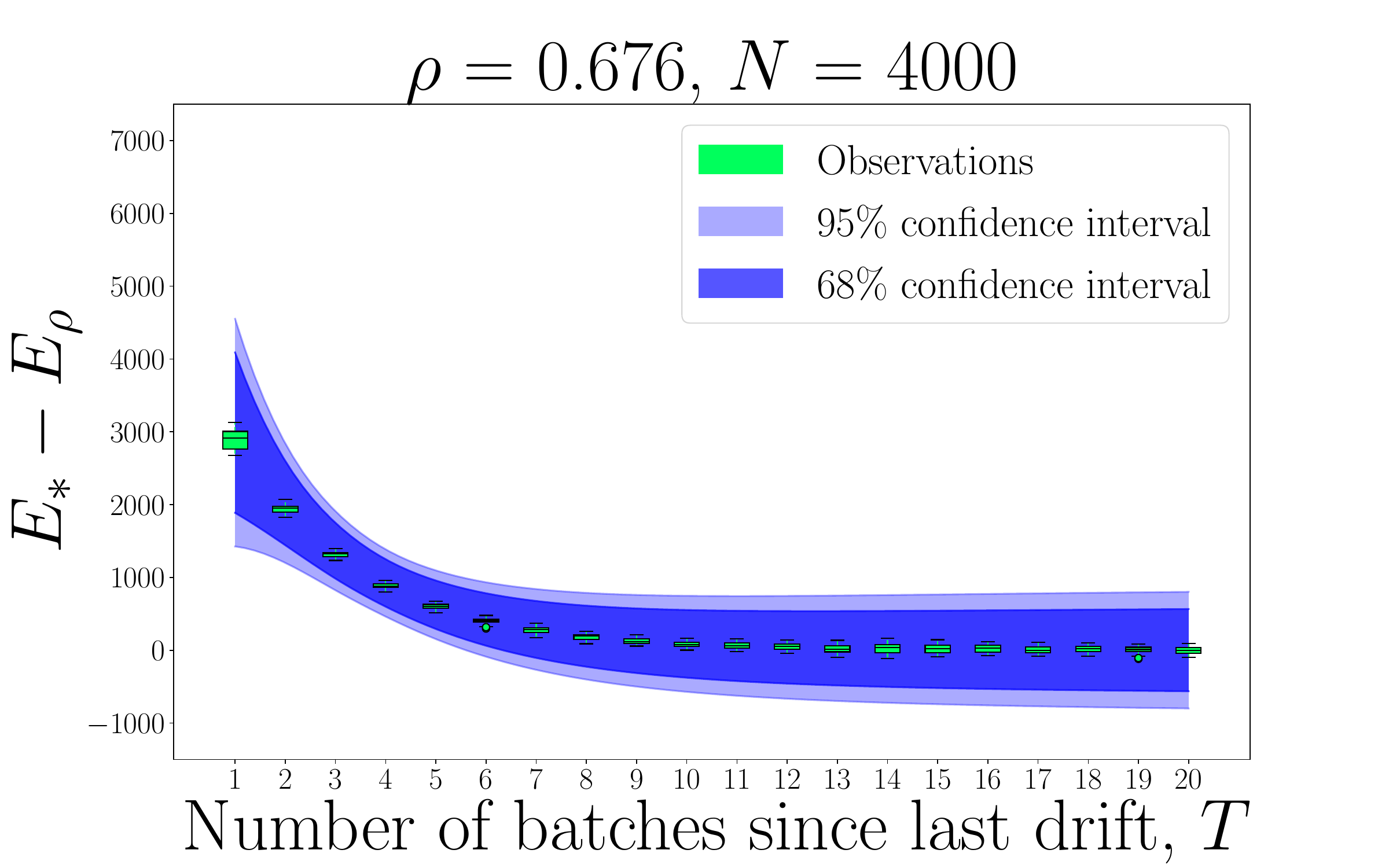}}
	\hfill
	\subfloat[]{\includegraphics[width=0.333\textwidth]{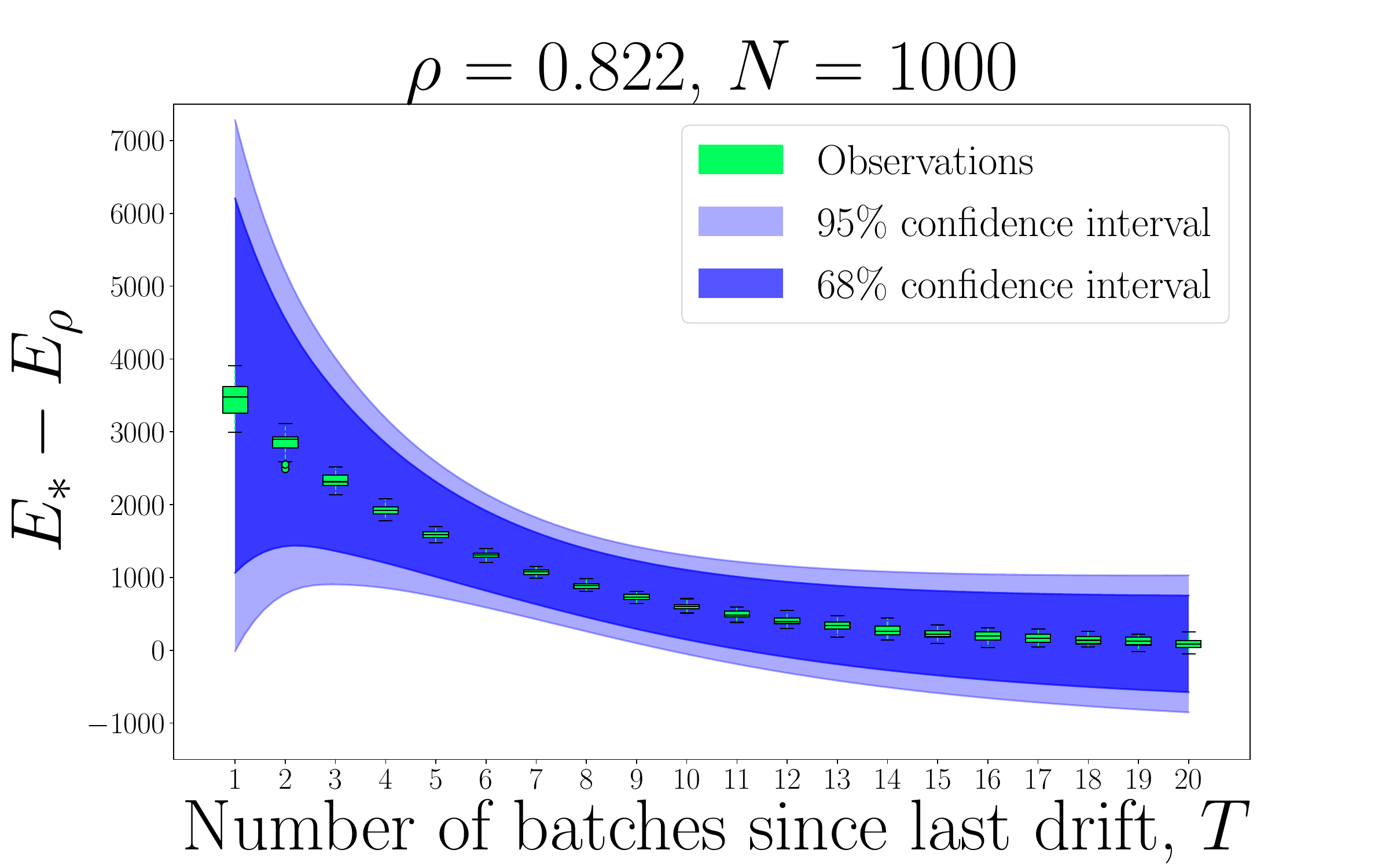}}

	\caption{Difference between the S$K$M error $E_*$ and the surrogate error $E_\rho$ as $T$ increases, for $\varepsilon=0.5$. }	
	\label{fig:Surrogate}
\end{figure*}

\begin{dfn} \label{Surrogate}
Given a set of batches of data points, $\mathcal{X}= \{B^t\}_{t\geq0}$, the surrogate error function, for a given set of centroids $C$, is defined as

\begin{equation}
   E_\rho(\mathcal{X},C)= \frac{1}{M_{\mathcal{X}}} \cdot \sum_{t\geq0} \rho^t \cdot \sum_{x \in B^t} \|\textbf{x}-\textbf{c}_x\|^2 
\end{equation}
where $M_{\mathcal{X}}=\sum_{t\geq0}\rho^t\cdot|B^t|$ is the total weighted mass of the set of batches $\mathcal{X}=\{B^t\}_{t\geq 0}$.
\end{dfn}

The surrogate error is a weighted version of the $K$-means error for SD. Furthermore, the following theorem illustrates the suitability of this alternative function. For the sake of simplicity, we consider for this theorem that all batches have the same number of data points, $|B^t|=N$.

\begin{restatable}{thm}{primteo} \label{thm:surrogate}
Let $\textbf{c} \in \mathbb{R}^d$ be a point, $\mathcal{X}=\{B^t\}_{t\geq 0}$ be a set of batches of points in $\mathbb{R}^d$, where $B^t=\{\textbf{x}_i^t\}_{i=1}^{N}$ and $t$ denotes the antiquity of $B^t$. Let the batches before the drift $\{B^t\}_{t > T-1}$ be i.i.d. according to the probability $p$, where $\mathbb{E}_{p}[\|\textbf{x}-\textbf{c}\|^2] = E$. Let the batches after the drift $\{B^t\}_{t\leq T-1}$ be i.i.d according to $p'$, where $\mathbb{E}_{p'}[\|\textbf{x}-\textbf{c}\|^2]=(1+\varepsilon)\cdot E$ for $\varepsilon > 0$. Let us assume that $\|\textbf{x}_i^t-\textbf{c}\|^2$ is upper-bounded by $u \geq 0$, for $i= 1, ..., N$ and $t \geq 0$.

Then, with at least probability $1-\delta$, the difference $E_{*}(\mathcal{X},\{\textbf{c}\})-E_\rho(\mathcal{X},\{\textbf{c}\})$ satisfies:
\begin{equation}\label{eq:Bounds}
E_{*}(\mathcal{X},\{\textbf{c}\})-E_\rho(\mathcal{X},\{\textbf{c}\}) \in (\rho^{T}\cdot \varepsilon \cdot E -\gamma , \rho^{T}\cdot \varepsilon \cdot E + \gamma),
\end{equation}
where
\begin{equation}\label{eq:ProbabilityConfidence}
\gamma= u \cdot \sqrt{\frac{(2\cdot \rho^{T} - 1)/T + (1 - \rho)/(1 + \rho)}{2\cdot N}\cdot \ln \frac{2}{\delta}},
\end{equation}

\end{restatable} 

 For this theorem we do not assume any specific underlying distribution, and the only assumption is that the squared distance with respect to $\textbf{c}$ is bounded by $u$. We present a $(1+\varepsilon)$-drift as well, which occurs when two underlying distributions $p$ and $p'$ satisfy $\mathbb{E}_{p'}[\|\textbf{x}-\textbf{c}\|^2]=(1+\varepsilon)\cdot\mathbb{E}_{p}[\|\textbf{x}-\textbf{c}\|^2]$, for $\varepsilon>0$. More importantly, observe that, according to Theorem \ref{thm:surrogate}, the expected value of the alternative error function tends to the S$K$M error function exponentially fast with $T$ for a single centroid, since the mean value of their difference has the form $\rho^{T}\cdot\varepsilon\cdot E$. In particular, it shows that the surrogate function can be used to approximate the error for a single centroid, thus applying this result to every subgroup of points and their respective centroids yields a good approximation of the S$K$M error. In summary, Theorem \ref{thm:surrogate} shows that  we can deal with the S$K$M problem by minimizing the alternative error without having to detect concept drifts. 
 
Due to the exponential decrease of the weights as antiquity $t$ increases, the contribution to the approximated error of older batches rapidly becomes negligible. Therefore, in practise, we can compute an arbitrarily accurate surrogate error function by considering the last $T_{max}$ batches. By using this approximation, we deal with the issue of indefinite increasing volume of data. 

In Figure \ref{fig:Surrogate}, we show how $E_*-E_\rho$ tends to zero as $T$ increases. For these experiments, we have stored 40 batches of size $N$ with a specific concept, and then 20 batches of a $(1+\varepsilon)$-drift were added sequentially. Here, the centroid $\textbf{c}$ was stated as the center of mass of the data points\footnote{The data points of both concepts were previously generated, and are chosen randomly for each batch.} from the first concept, and $u$ is the distance from the farthest point to the centroid $\textbf{c}$. For each new batch $T$ increases by 1, and we compute the difference between both errors (S$K$M and surrogate) and their theoretical bounds $\gamma$ (\ref{eq:ProbabilityConfidence}). Because the theorem gives a probabilistic result, we have repeated the experiment many times, randomly selecting batches at each run. Figure \ref{fig:Surrogate} shows the computed differences with a boxplot layout since this experiment was repeated 20 times in order to obtain statistical results. Two confidence intervals are given in the figure, with probabilities 95\% and 68\%, which correspond to values of $\delta=$ 0.05 and 0.32, respectively.

A batch size of $N=1000$ was set in (a) and $N=4000$ in (b). Notice that $\gamma\propto\frac{1}{\sqrt{N}}$, therefore, as the number of points on each batch increases, the bounds become narrower. Additionally, $\rho$ was set equal to $0.676$ for (a) and (b) and $0.822$ for (c). On one hand, lower values of $\rho$ makes the expected difference between the S$K$M error and the approximation tends to zero faster. In other words, the bias of the surrogate as an estimate of the S$K$M error decreases faster for smaller values of $\rho$. On the other hand, lower values of $\rho$ implies broader bounds of $E_*-E_\rho$. Thus, the variance of the surrogate estimate is higher as $\rho$ decreases. Clearly, there is a trade-off between fast convergence and low variance when choosing the forgetting parameter $\rho$. Hence it should be chosen attending the requirements of the user; we propose a heuristic to set the $\rho$ value in section \ref{sec:expSetup}.

\section{Streaming Lloyd's Algorithm}\label{sec:Initialization}

In order to deal with the S$K$M problem in a passive way, we propose the Forgetful S$K$M (FS$K$M) algorithm (Algorithm \ref{FSKMa}). FS$K$M approximates the solution of the S$K$M problem by minimizing the approximated error function. When a new batch $B^0$ arrives, FS$K$M runs an initialization procedure to find a set of initial centroids. Next, a weighted Lloyd's algorithm is carried out over the available set of batches $\mathcal{X}$. The running time of the weighted Lloyd's algorithm is  $\mathcal{O}(n\cdot K\cdot d)$, where $n$ is the total number of points to be clustered. However, recall that we can compute an arbitrarily accurate surrogate error function by discarding batches with a negligible weight. By discarding the batches with negligible weights, the computational complexity of the weighted Lloyd's step of FS$K$M is reduced to $\mathcal{O}(M_{T_{\max}} \cdot K\cdot d)$, where $M_{T_{max}}$ is the sum of batch sizes of the stored batches. This is a passive approach, because it does not need to detect concept drifts, and it inherits the good properties of the surrogate function as alternative to the real S$K$M error.

\begin{algorithm}[!t]
\caption{Forgetful S$K$M algorithm (FS$K$M)}\label{FSKMa}
\begin{algorithmic}[1]
\State \textbf{Predetermined:} Maximum number of batches saved $T_{max}$ and forget parameter $\rho$.
\State \textbf{Input:} Set of previous batches $\mathcal{X}$ partitioned by $\mathcal{P}$ and new batch $B^0$.
\State \textbf{Output:} A set of centroids $C$ and its associated partition $\mathcal{P}$.

\If{size($\mathcal{X}$)==$T_{max}$}
\State $\mbox{Remove the oldest batch from }\mathcal{X}$
\EndIf
\State $\mathcal{X} \gets \mbox{Append }\ B^0$
\State $C \gets \mbox{Initialization}(\mathcal{X},\mathcal{P})$
\State $C,\mathcal{P} \gets \mbox{Weighted Lloyd}(\mathcal{X},C)$
\State\Return $C,\mathcal{P}$
\end{algorithmic}
\end{algorithm}

\subsection{Initialization Step}

As previously discussed, the initialization stage has a major effect on the convergence of Lloyd's, and therefore of the FS$K$M algorithm. For this reason, in this section, we propose efficient procedures for the initialization step of FS$K$M. However, first we describe two simple approaches, which will be used for comparison. Once a new batch is received, a straightforward initialization strategy is to use the previously converged set of centroids. We call this approach Previous Centroids (PC), and the set of centroids obtained in previous iterations will be denoted as $C^*=\{\mathbf{c}_k^*\}_{k=1}^{K}$. PC uses a set of locally optimal centroids for the past set of batches, which can be a good and efficient choice once a new batch is presented. An alternative straightforward initialization is to use centroids obtained by applying a standard initialization procedure over the newest batch $B^0$, such as KM++. We call this approach the Current Centroids (CC). The set of centroids obtained from initializing over the current batch is denoted as $C^0=\{\mathbf{c}_1^0\}_{k=1}^{K}$. Clearly, CC allows FS$K$M to adapt rapidly when a concept drift occurs. However, this initialization does not take into account either the batches from the past or the set $C^*$. This could imply a waste of very valuable information, especially when a concept drift has not occurred for a long period of time.

\subsection{Weighted Initialization}
Considering the trade-off between the PC and CC approaches, we propose two efficient initialization strategies that combine information from PC and CC, by minimizing an upper-bound to the surrogate error function. The next result defines an upper-bound for the surrogate error function that will allow us to determine a competitive initialization for the FS$K$M algorithm.

\begin{restatable}{thm}{secteo}
\label{theo:bound}
Given two sets of centroids $C^*=\{\mathbf{c}_k^*\}_{k=1}^{K}$ and $C^0=\{\mathbf{c}_k^0\}_{k=1}^{K}$, for any set of centroids $C \in \mathbb{R}^d$, the surrogate function $E_\rho(\mathcal{X},C)$ can be upper-bounded as follows:
\begin{align}\label{SurrogateBound2}
E_\rho(\mathcal{X},C)\leq f^\rho(\mathcal{X},C) + const, 
\end{align} 
where 
\begin{align}
\label{eq:generalUpper}
f^\rho(\mathcal{X},C)\equiv& \frac{1}{M_{\mathcal{X}}}\cdot \sum_{k=1}^{K}\big(w^*_k \cdot \|\textbf{c}_{k'}-\textbf{c}^*_k\|^2+w^0_k\cdot \|\textbf{c}_{k''}-\textbf{c}^0_k\|^2\big),
\end{align}
for $\textbf{c}_{k'}= \underset{\textbf{c}\in C}{\arg\min}\|\textbf{c}^*_k-\textbf{c}\|$, $\textbf{c}_{k''}=\underset{\textbf{c}\in C}{\arg\min}\|\textbf{c}^0_k-\textbf{c}\|$, where $w^*_k=\sum_{t\geq1}^{} \rho^t \cdot |B^{t}\cap P_{k}^{*}|$ and $w^0_k=|B^{0}\cap P_{k}^{0}|$ are the weights related to each centroid
and $const$ is a value independent of the set of centroids $C$. $B^{t}\cap P_{k}$ are the set of points in batch $B^t$ that belong to the partition $P_{k}$.
\end{restatable}

Theorem \ref{theo:bound} shows that the surrogate error is upper-bounded by
$f^\rho(\mathcal{X},C)$ plus a constant. In fact, observe that $f^\rho$ has the form of a weighted $K$-means error with $\{\textbf{c}^*_k,\textbf{c}^0_k\}_{k=1}^{K}$ as the data points, and weights $W=\{w^*_k,w^0_k\}_{k=1}^{K}$. Hence, we propose an initialization procedure based on the weighted $K$-means algorithm over the union of both sets of centroids. We refer to this initialization as Weighted Initialization (WI, Algorithm \ref{algo:Initializationa}), where its computational complexity is $\mathcal{O}(K \cdot \max\{|B^0|, K\}\cdot d)$.

\begin{algorithm}[!t]
\caption{Weighted $K$-means initialization (WI)) }\label{algo:Initializationa}
\begin{algorithmic}[1]
\State \textbf{Predetermined:} Number of clusters $K$. Forgetting parameter $\rho$.
\State \textbf{Input:} A set of batches $\mathcal{X}=\{B^t\}_{t\geq 0}$, a set of previous centroids $C^*$ which are induced by the partition $\mathcal{P}^*$.
\State \textbf{Output:} A set of new optimized centroids $C$ and its associated partition $\mathcal{P}$.

\State $C^0,\mathcal{P}^0 \gets \mbox{KM++}(B^0)$
\State $w^*_k,w^0_k \gets \mbox{Compute weights from }\mathcal{P}^* \mbox{ and } \mathcal{P}^0$
\State $C,\mathcal{P} \gets \mbox{Weighted $K$-means}(X=\{C^*,C^0\},W=\{w^*_k,w^0_k\}_{k=1}^{K})$
\State\Return $C,\mathcal{P}$
\end{algorithmic}
\end{algorithm}
  
\subsection{Hungarian Initialization}

An interesting analytical result can be acquired considering another assumption along with Theorem \ref{theo:bound}.  Assume that each centroid $\textbf{c}_k$ has a single pair of centroids $\textbf{c}_k^*,\textbf{c}_{\sig}^0$ which are the closest to itself from both sets $C^*$ and $C^0$, and are distinct for each centroid $\textbf{c}_k$. We can index the centroid $\textbf{c}_k$ like the centroids in $C^*$, but a different indexation $k'=\sig$ may be needed for the centroids in $C^0$, represented by the permutation $\sig$. Then, we can re-write the upper-bound given in (\ref{eq:generalUpper}) as follows: 
\begin{align}\label{SimpleUpperBound}
f^\rho(\mathcal{X},C=\{\textbf{c}_k\}_{k=1}^{K})&=\frac{1}{M_{\mathcal{X}}}\cdot \sum_{k=1}^{K}\big(w^*_k\cdot \|\textbf{c}_{k}-\textbf{c}^*_k\|^2+\nonumber\\
&+w^0_{\sigma(k)}\cdot \|\textbf{c}_{k}-\textbf{c}^0_{\sig}\|^2\big),
\end{align}
where the weights $w^*_k$ and $w^0_{\sig}$ are the weights of $c^*_k$ and $c^0_{\sig}$, respectively. The following theoretical result shows that the upper-bound $f^\rho(\mathcal{X},C)$ can be analytically minimized with respect to $\textbf{c}_k$ with this assumption.

\begin{restatable}{thm}{quadteo}\label{thm:MinimizeSurrogate}
Let $f^\rho(\mathcal{X},C)$ be the function defined in (\ref{SimpleUpperBound}) for a set of centroids $C=\{\textbf{c}_k\}_{k=1}^{K}$ of size $K$, where $\textbf{c}^*_k$ and $\textbf{c}^0_{\sig}$ are given, and they are the closest points to $\textbf{c}_k$ of the sets $\{\textbf{c}^*_k\}_{k=1}^{K}$ and $\{\textbf{c}^0_{\sig}\}_{k=1}^{K}$. Then the set of centroids that minimizes this function is given by:
\begin{align}
\label{eq:HungarianInit}
\textbf{c}_k = \frac{1}{w^*_k+w^0_{\sig}}\cdot(w^*_k\cdot\textbf{c}^*_k+w^0_{\sig}\cdot\textbf{c}^0_{\sig}),
\end{align}
\end{restatable}

Theorem \ref{thm:MinimizeSurrogate} shows that, just by making the one-to-one assumption given by $\sigma$, the optimal centroids $C$ can be simply expressed as a linear combination between the elements of $C^*$ and $C^0$. Notice that with this assumption we achieve an analytical minimum of $f^\rho(\mathcal{X},C)$.

\paragraph{Linear Sum Assignment Problem}\label{sec:linearsum} If we want to compute the optimal centroids under the previous assumption, $\sig$ must be found. In order to do so, we use the result in Theorem \ref{thm:MinimizeSurrogate} to rewrite (\ref{SimpleUpperBound}):  

\begin{align}
f^\rho(\mathcal{X},C)=\frac{1}{M_{\mathcal{X}}}\cdot \sum_{k=1}^{K}\frac{w^*_k\cdot w^0_{\sig}}{w^*_k+w^0_{\sig}}\cdot\|\textbf{c}^*_k-\textbf{c}^0_{\sig}\|^2
\end{align}

Hence, we define the matrix:

\begin{align}
f_{k,k'}=\frac{w^*_k\cdot w^0_{k'}}{w^*_k+w^0_{k'}}\cdot \|\textbf{c}^*_k-\textbf{c}^0_{k'}\|^2 \ ; \ k,k'\in \{1,\ldots, K\},
\end{align}

\noindent and find the permutation $\sigma$ such that the sum $\sum_{k=1}^K f_{k,\sig}$ is minimal. This is a linear sum assignment problem and we can make use of the Hungarian (or Kuhn-Munkres) algorithm \cite{Hungarian} to determine $\sig$ with a computational complexity of $\mathcal{O}(K^3)$. Thus, we propose another initialization method called Hungarian Initialization (HI) based on this procedure. HI firstly computes a set of optimized centroids $\textbf{c}^0_{k'}$ over the new batch $B^0$. Then the matrix $f_{k,k'}$ is constructed, which is used to determine the permutation that maps $k\rightarrow k'=\sigma(k)$, via the linear sum assignment problem (Algorithm \ref{algo:Initializationa2a}). This way, the sum $\sum_{k=1}^K f_{k,\sigma(k)}$ is guaranteed to be the minimum value of $f^\rho(\mathcal{X},C)$, and hence the new set of centroids can be computed as defined in Theorem \ref{thm:MinimizeSurrogate}. The computational complexity of this algorithm is $\mathcal{O}(K\cdot \max\{\max\{K^2,K\cdot d\},|B^0|\cdot d\})$.

\begin{algorithm}[!t]
\caption{Hungarian Initialization (HI)}\label{algo:Initializationa2a}
\begin{algorithmic}[1]
\State \textbf{Predetermined:} Number of clusters $K$. Forgetting parameter $\rho$.
\State \textbf{Input:} A set of batches $\mathcal{X}$, ordered in such a way that $B^0$ is the newest batch. A set of previous centroids $C^*$.
\State \textbf{Output:} A set of new optimized centroids $C$.

\State $C^0 \gets \mbox{KM++}(B^0)$
\For{k in 1,...,K}
\For{k' in 1,...,K}
\State $f_{k,k'}\gets \frac{w^*_k\cdot w^0_{k'}}{w^*_k+w^0_{k'}}\cdot \|\textbf{c}^*_k-\textbf{c}^0_{k'}\|^2$
\EndFor
\EndFor
    \State $\sigma\gets \underset{\sigma\in \Sigma}{\arg\min} \sum_{k=1}^K f_{k,\sigma(k)}$
\State $C\gets\emptyset$
\For{k in 1,...,K}
\State $\textbf{c}_k\gets\frac{1}{w^*_k+w^0_{\sigma(k)}}\cdot(w^*_k\cdot \textbf{c}^*_k+w^0_{\sigma(k)}\cdot \textbf{c}^0_{\sigma(k)})$
\State $C\gets C\cup\textbf{c}_k$
\EndFor
\State\Return $C$
\end{algorithmic}
\end{algorithm}

\section{Experiments}\label{Experiments}

In this section we analyse the performance of the FS$K$M algorithm with the proposed initialization procedures: Previous Centroids (PC), Current Centroids (CC), Hungarian Initialization (HI) and Weighted Initialization (WI). The converged S$K$M error obtained by FS$K$M with different initialization strategies is compared with the gold-standard PS$K$M. In order to control the strength of the drifts, the experiments are performed using simulated streaming data with $(1+\varepsilon)$-drifts generated using real datasets taken from the \textit{UCI Machine Learning Repository} \cite{UCI}, for different values of $\varepsilon$. For further insight on how we simulated streaming data see Appendix \ref{app:sim}.


\subsection{Experimental Setup}\label{sec:expSetup}

\paragraph{Datasets} The experiments have been carried out in 8 different datasets simulated based on real datasets from the \textit{UCI Machine Learning Repository} \cite{UCI}, see Table \ref{tab:Datasets}.  The simulated data consists of a sequence of batches with size $N=500$, and a $(1+\varepsilon)$-concept drift takes place every 10 batches.

\begin{table}[!t]
\centering
\begin{tabular}{|c|c|c|}
\cline{1-3}
 Dataset& $d$& $n$  \\ \hline
\multicolumn{1}{|l|}{\textbf{Urban accidents}\cite{Urban} }& 2 & $3.6\cdot 10^5$  \\ \hline
\multicolumn{1}{|l|}{\textbf{Pulsar detection}\cite{Pulsar}} & 8 &  $1.8\cdot 10^4$  \\ \hline

\multicolumn{1}{|l|}{\textbf{SUSY}\cite{SUSY}} & 18 & $5\cdot 10^6$   \\ \hline

\multicolumn{1}{|l|}{\textbf{Gas sensors}\cite{Gas1,Gas2}} & 20 & $4\cdot 10^6$   \\ \hline
\multicolumn{1}{|l|}{\textbf{Anuran calls}\cite{Frogs}} & 22 &  $7.2\cdot 10^3$
\\ \hline
\multicolumn{1}{|l|}{\textbf{Google Reviews}\cite{Google}} & 25 &  $5.5\cdot 10^3$
\\ \hline
\multicolumn{1}{|l|}{\textbf{Gesture Segmentation}\cite{Gesture}} & 50 & $9.9\cdot 10^3$
\\ \hline
\multicolumn{1}{|l|}{\textbf{Epilepsia}\cite{Epilectic}} & 178 &  $1.1\cdot 10^4$
\\ \hline
\end{tabular}

\caption{Information of the considered datasets.}
\label{tab:Datasets}
\end{table}

\paragraph{Procedure} To analyze the behavior of the algorithms in streaming scenarios, we perform a burning-out step by storing $T_{max}$ batches from the first concept. After this step, we measure the evolution of the performance of PS$K$M, and FS$K$M with different initialization techniques. To fairly compare their behaviour, the set of centroids $C^*$ and $C^0$ are the same for each initialization procedure each time a new batch arrives. After the burning-out step, a stream of 100 batches is processed with concept drifts every 10 batches. This procedure is repeated for each dataset and value of the hyperparameters. Because the results did not vary too much for intermediate batches, we show measurements for the first and second batch (indexed by $1$ and $2$), an intermediate batch and the last batch before the next concept drift (indexed by $4$ and $10$). 

\paragraph{Measurements} We have measured the quality of the solutions obtained by different procedures in terms of the S$K$M and approximated error function. In order to have comparable scores for different datasets, the obtained scores (error values) on initialization and convergence errors are normalized. For each new batch, the score $E_M$ obtained with algorithm $M \in \mathcal{M}$ is normalized with respect to the minimum over every algorithm $\mathcal{M}$ as $\hat{E}_M=(E_M-\underset{M'\in\mathcal{M}}{\min}E_{M'})/\underset{M'\in\mathcal{M}}{\min}E_{M'}$. Using normalized scores allows to summarize the results obtained for different algorithms with all the data sets in a single plot, reducing dramatically the number of figures needed to display the results. The computational load of the methods considered in our experimental setting is dominated by the number of distance computations. Therefore,
as it is common practice in $K$-means problem related articles \cite{kmc2,Capo1}, we use the number of distances computed to measure their computational performance. The computed distances $D_M$ were also normalized, but  we simply divide by the minimum $\hat{D}_M=D_M/\underset{M'\in\mathcal{M}}{\min}D_{M'}$. This way, in the figures of Section \ref{sec:resultstime} what will be shown is how many times more distances have been computed compared to the fastest one. 

\paragraph{Hyperparameters} A key parameter is the forget parameter $\rho$, since the approximated function directly depends on this parameter. Theorem \ref{thm:surrogate} shows that the surrogate differs from the real S$K$M error with $\rho^T\varepsilon$, but the confidence interval grows as $\rho$ decreases, hence a proper balance is necessary. Assuming that a difference of $0.01$ is negligible, we can determine the $\rho$ value by solving the equality $\varepsilon\cdot\rho^{\tau/m}=0.01$, where $\tau$ is our prior knowledge about the (average) number of batches in which a concept is stable and $m$ represents the fraction of the period in which we want the difference to become negligible. Intuitively, $m$ determines how fast the term $\rho^T\cdot\varepsilon$ shrinks relative to the period of when a drifts happens, $\tau$. The magnitude of the concept drift and the number of clusters can affect how fast each algorithm adapts. For this reason, when generating streaming data, we use the next set of values for the parameters $\varepsilon$ and $K$: $\varepsilon \in \{0.5,1,2\}$, $K \in \{5,10,25\}$. Note that for each value of $\varepsilon$ and $m$, we set a different value of $\rho$. In this paper, we show results for $m=2$, for the sake of brevity. For $m=2$, the values of $\rho$ depended on $\varepsilon$, which were $\rho\in\{0.457,0.398,0.347\}$ in the order of $\varepsilon$ given before.

\subsection{Initial and Converged Errors}\label{sec:resultserror}

\begin{figure}[!t]
	\centering
	\subfloat{\includegraphics[width=0.5\textwidth]{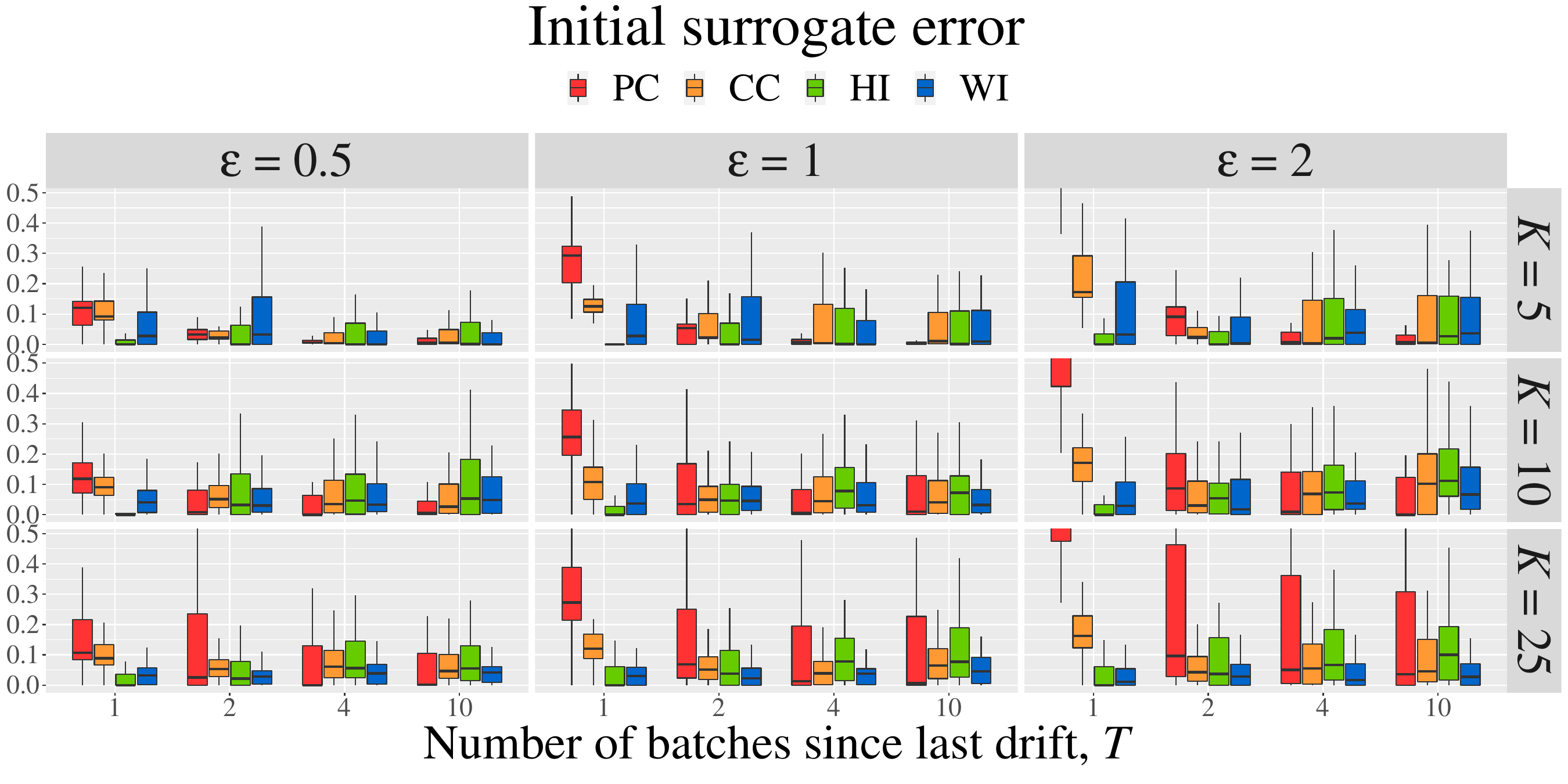}}
	\caption{Initial surrogate error for the FS$K$M algorithm with different initialization methods. $\rho$ was determined with $m=2$, and errors where normalized as specified in the former section.}
	\label{fig:Initial error}
\end{figure}

 HI and WI show better initial surrogate errors compared with PC and CC when a concept drift occurs (see Figure \ref{fig:Initial error}, $T=1$), for every $\varepsilon$ and $K$. When a concept drift occurs, PC performs poorly, since its initial centroids are focused on minimizing the approximated error function for the previous concept.  For smaller values of $\rho$, CC gets better results than PC when a drift occurs, because previous batches contribute less to the surrogate error. In this sense, CC gets better results than PC as $\varepsilon$ increases, because previously computed centroids become an even worse approximation for the novel concept. As new batches arrive, we observe that PC obtains the best initial surrogate error, because stored batches share the same underlying distribution and previously converged centroids $C^*$ are a good initialization.

\begin{figure}[!t]
	\centering
	\subfloat{\includegraphics[width=0.5\textwidth]{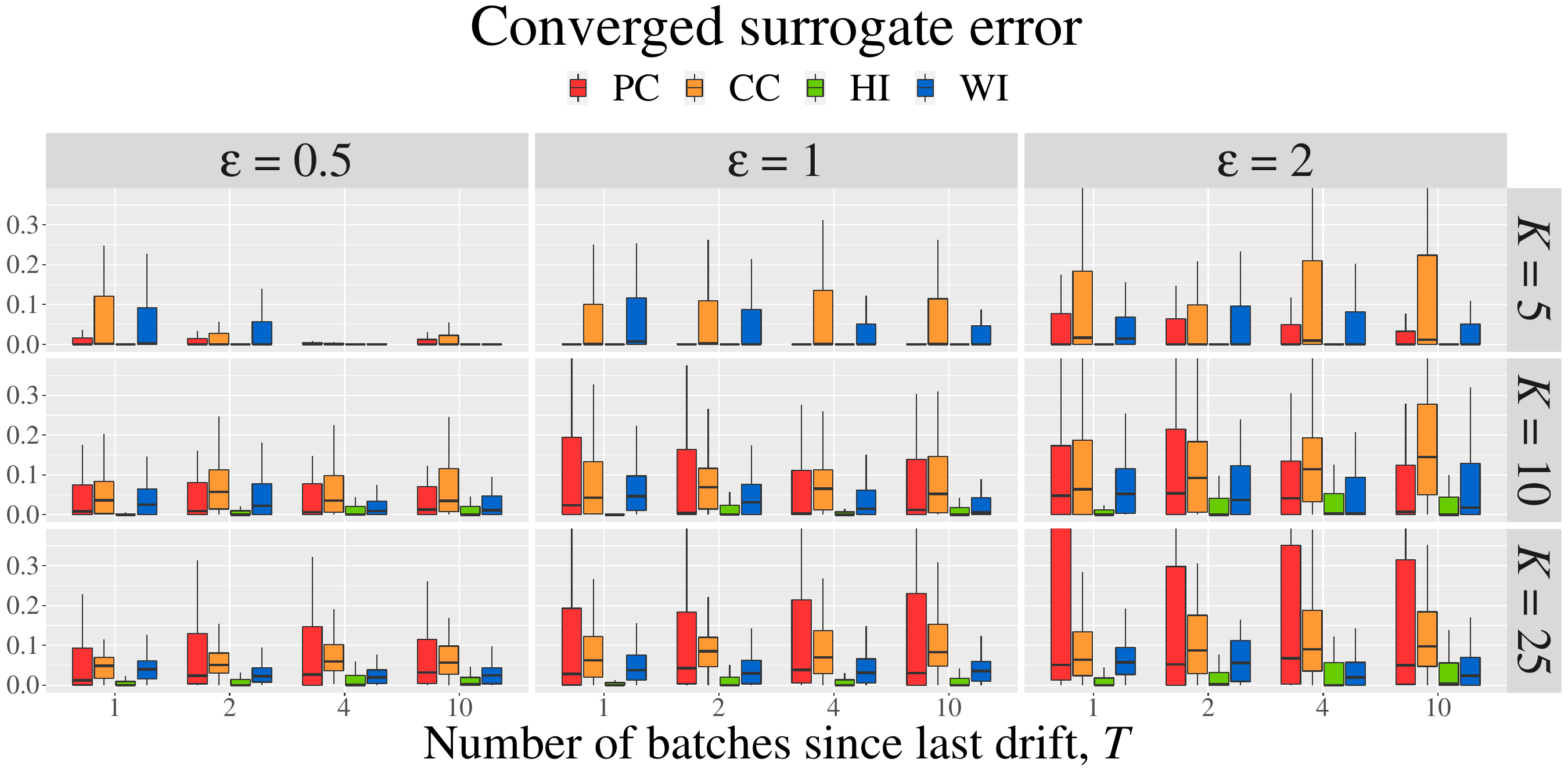}}
	\caption{Converged surrogate error for the FS$K$M algorithm with different initialization methods. $\rho$ was determined with $m=2$, and errors where normalized.}
	\label{fig:Initial real error}
\end{figure}

Figure \ref{fig:Initial real error} summarizes the surrogate error function of FS$K$M at convergence. HI and WI stand out over the trivial initialization methods. Moreover, HI obtains median scores close to $0$ for every value of $K$ and $\varepsilon$. In the previous figure, it was shown that WI obtained a better initialization error, but now HI obtains a lower converged error. HI initialization is more restricted than WI, obtaining a worse initialization error. However, this restriction seems to be reasonable since the fixed points where HI arrives get a better converged error. Furthermore, WI executes $K$-means over centroids, and completely ignores the structure of data points, which may lead to re-assignations that increase the error. PC shows a higher variance, especially for bigger values of $K$.

\begin{figure}[!t]
	\centering
	\subfloat{\includegraphics[width=0.5\textwidth]{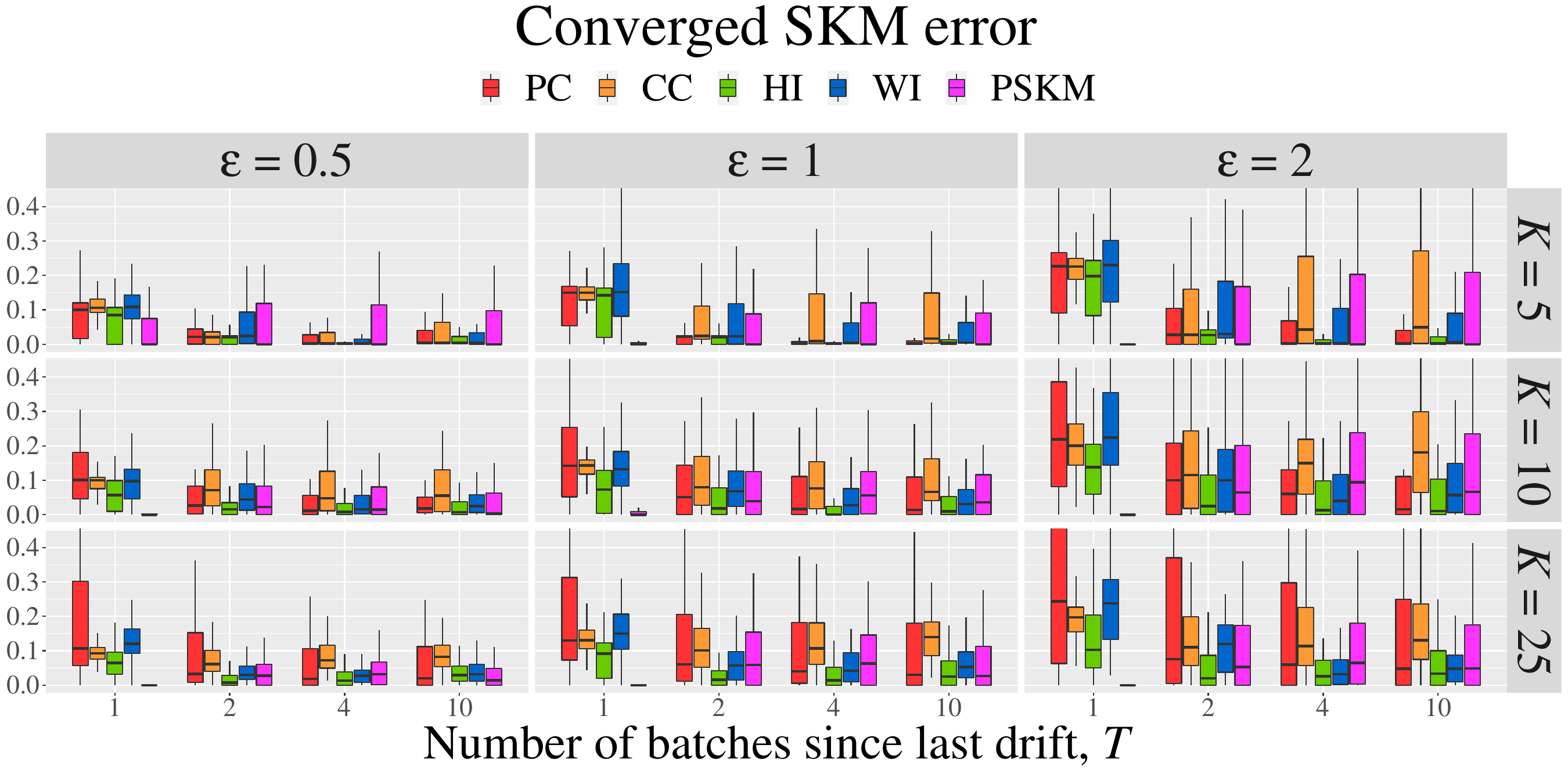}}
	\caption{Converged real S$K$M error for the FS$K$M (multiple initialization) and PS$K$M algorithms. Here $m=2$ and error were normalized.}
	\label{fig:Converged real error}
\end{figure}

In Figure \ref{fig:Converged real error}, we show the S$K$M error at convergence. Here the results of PS$K$M are shown as a reference. Observe that, in general, the medians of the converged S$K$M error are comparable for every algorithm, especially when many batches of the same concept have already happened ($T=10$). Recall that FS$K$M does not minimize the S$K$M error, concluding that the surrogate is a good approximation and that every initialization technique (except for CC) works properly. We see that even though PS$K$M obtains the best scores when a drift occurs, after the next batch (index $2$) HI and WI already attain scores comparable to PS$K$M in terms of medians. In terms of dispersion, HI and WI are even more stable (smaller variance) than PS$K$M. We know from Theorem \ref{thm:surrogate} that the surrogate error approximates the S$K$M error better when more batches occurred since the last concept drift, this can explain why, even though FS$K$M does not explicitly minimize the S$K$M error, its convergence value is better than the one obtained by PS$K$M. We can see that in the last batch, before a concept drift occurs, FS$K$M obtains scores comparable to PS$K$M.

\begin{figure}[!t]
	\centering
	\subfloat{\includegraphics[width=0.5\textwidth]{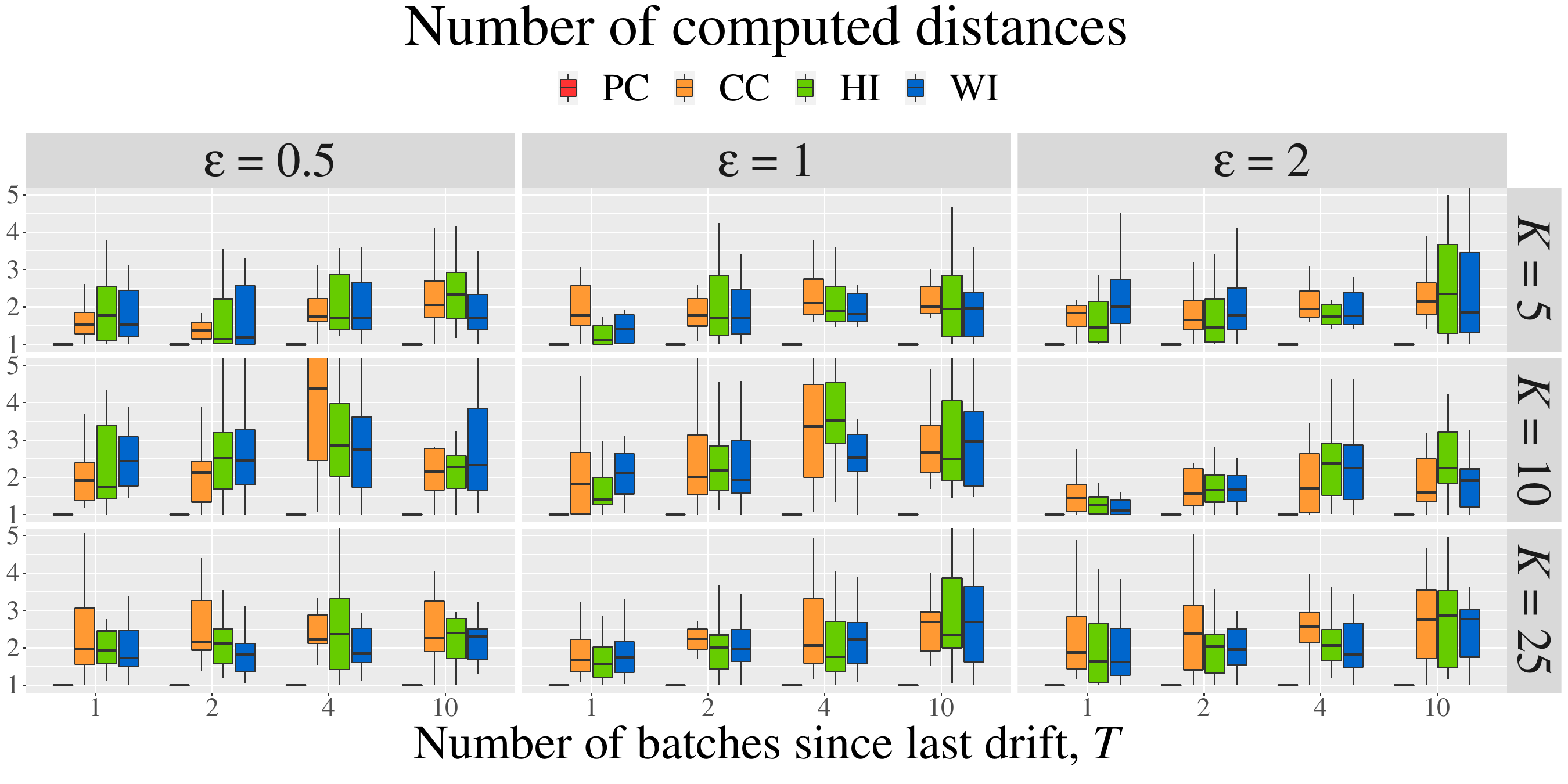}}

	\caption{Number of computed distances, normalized as $\hat{D}_M=D_M/\underset{M'\in\mathcal{M}}{\min}(D_{M'})$. PC's boxplot is flat as it is always the procedure which computes the less number of distances, since its initialization needs no computation. }
	\label{fig:ndis}
	
\end{figure}

\subsection{Computed Distances}\label{sec:resultstime}

Not needing any extra computation for the initialization makes PC compute fewer distances, thus we use PC as a reference in Figure \ref{fig:ndis}, where the number of distances is shown relative to PC's number of computed distances. Because distances are normalized divided by the minimum obtained over every algorithm, what we observe in the Y axis is how many times more distances have been computed compared to PC. Considering every boxplot, we conclude that the medians of HI, WI and CC are around $2$, thus they compute twice as many distances as PC in general. However, in the previous section we have observed that HI outperformed in terms of converged S$K$M error, thus this extra distance computation is a trade-off we must pay in order to adapt to concept drifts more effectively.

\section{Conclusions}

In this work we have proposed an approximation function for the S$K$M error, which can be computed without requiring the concept drift detection.  We have proved that the surrogate is a good approximation to the S$K$M error, and its quality improves as the number of batches from the same concept increases. 


We have performed a set of experiments using real data as a basis and simulated streaming scenarios with $(1+\varepsilon)$ concept drifts. We have compared minimizing the approximated error to minimizing the actual S$K$M error. The behavior of minimizing the surrogate has been analyzed for the proposed initialization procedures. In the last section, we have seen that the proposed initialization algorithms stood out over the trivial methods, except in the number of computed distances, which was expected. Using previously computed centroids proves to be the fastest method, although it performs badly when a drift occurs. Because every other initialization method requires more steps, they need more iterations, which implies more computed distances and hence bigger elapsed time. However, this is a trade-off in exchange for a better response to concept drifts, more stable solutions and smaller error values, which is the main interest in the Streaming $K$-means problem. 

\section*{Acknowledgments}

This research is supported by the Basque Government through the BERC 2018-2021 program and by Spanish Ministry of Sciences, Innovation and Universities: BCAM Severo Ochoa accreditation SEV-2017-0718.

\appendices

\section{Simulated Streaming Data with $(1+\varepsilon)$-drifts}\label{app:sim}

Here we propose a heuristic method to generate an artificial SD with controlled concept $(1+\varepsilon)$-drifts based on a real dataset. However, first of all we shall recall what a $(1+\varepsilon)$-drift is. Let $C=\{\textbf{c}_k\}_{k=1}^{K}$ be a set of cluster centroids, and $X_1$ and $X_2$ be two sets of points with distributions $p$ and $p'$ respectively. Lets assume that the $K$-means error with respect to $C$ is bigger for $X_2$, then there is an $\varepsilon>0$ that satisfies $E(X_2,C)=(1+\varepsilon)\cdot E(X_1,C)$.

In order to generate controlled concept drifts (for a given $\varepsilon$ value), we start taking the real dataset $X_1$. This batch of data follows the distribution $p_1$, so our main objective is to obtain batches with different distributions $p_i$ such that (\ref{eq:ConceptDriftCondition sec:ConceptDriftGeneration}) is fulfilled for $i=1,...,M-1$, where $C_{i-1}$ are the clustering centroids computed for the $(i-1)$-th batch.

\begin{equation}\label{eq:ConceptDriftCondition sec:ConceptDriftGeneration}
    E(X_{i},C_{i-1})=(1+\varepsilon)\cdot E(X_{i-1},C_{i-1})
\end{equation}

\begin{figure}[!t]
    \centering
    \includegraphics[clip, trim=2cm 18cm 0 2cm, width=0.5\textwidth]{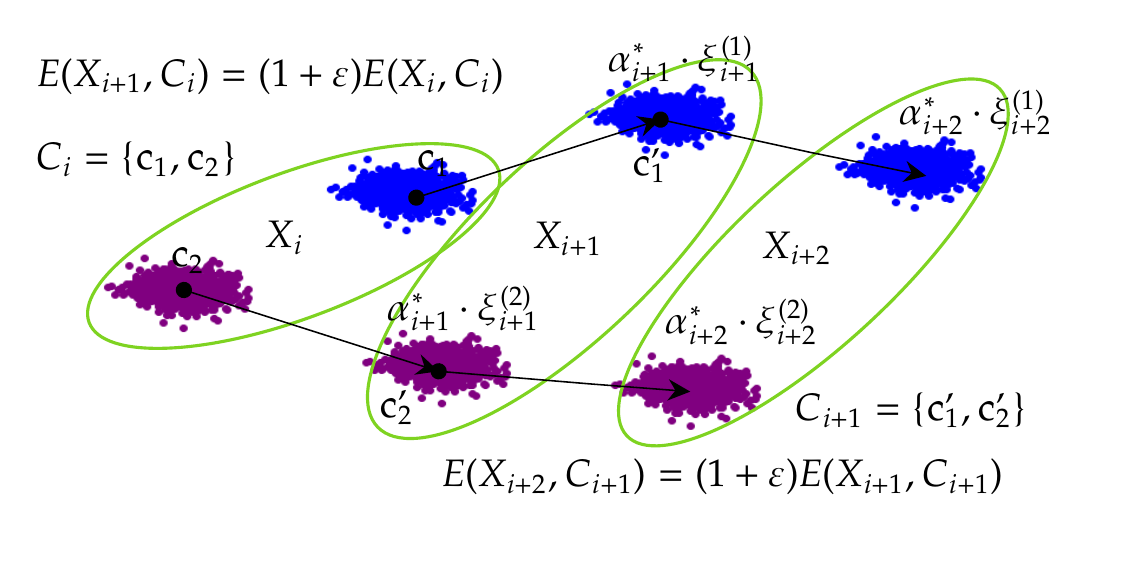}
    \caption{$(1+\varepsilon)$-drift generation. Here we observe a $K=2$ cluster set up in two dimensions, where $X_i$ is the first chunk of the example. And the following blocks are obtained following (\ref{translation}). Recall that each cluster is relocated with different random directions.}
    \label{CDGeneration}
\end{figure}

 The distribution can be changed by moving every point of each cluster in random directions. We generate new batches translating every point on each cluster $k$ in different random directions, however, the magnitude of the translation must be found heuristically such that (\ref{eq:ConceptDriftCondition sec:ConceptDriftGeneration}) is satisfied for distributions $p_i$ and $p_{i-1}$. We define $X_i$ as the data batch that follows the distribution $p_i$, where $X_1$ is the original data block. Every new block $X_i$ is obtained adding the vector $\alpha_i^*\cdot\mbox{\boldmath$\xi$}_i^{(k)}$ to every data point of cluster $k$ in $X_{i-1}$. Where $\mbox{\boldmath$\xi$}_i^{(k)}$ is a random unitary vector generated for each cluster and $\alpha_i^*$ is the optimal value of the magnitude obtained by the heuristic algorithm, we explain in the next section how to obtain it. We define $X_{i}^{(k)}$ to be the data points of cluster $k$ of $X^{(k)}_i$, and hence we generate the translated data as follows (See Figure \ref{CDGeneration} for illustration):
 
\begin{equation}\label{translation}
    X^{(k)}_{i} = X^{(k)}_{i-1}+\alpha_i^*\cdot\mbox{\boldmath$\xi$}_i^{(k)} 
\end{equation}

\subsection{Heuristic translation}

With this algorithm, a fit value for $\alpha_i^*$ is calculated so that (\ref{eq:ConceptDriftCondition sec:ConceptDriftGeneration}) is fulfilled, with a certain threshold. We start assigning an initial value $\alpha_i^1$. Since the error must increase $\varepsilon E(X_{i-1},C_{i-1})$, we assume that it must be of order:

\begin{equation}
    \alpha_i^1= \sqrt{\frac{\varepsilon E(X_{i-1},C_{i-1})}{K\cdot|X_{i}|}}
\end{equation} 

Hence we use the initial value $\alpha_i^{1}$, and we iteratively update $\alpha_i^{j}$ following (\ref{eq:alphaupdate sec:CDgeneration}) until we reach convergence. Here we define the error for each iteration $E_j$ as the one obtained translating the current data chunk with magnitude $\alpha_i^j$, $E_j\equiv E(X_{i-1}+\alpha_i^j\cdot\mbox{\boldmath$\xi$}_i,C_{i-1})$.

\begin{equation}\label{eq:alphaupdate sec:CDgeneration}
    \alpha_i^{j+1}=\alpha_i^{j}-\big(\frac{E_j}{(1+\varepsilon)\cdot E(X_{i-1},C_{i-1})}-1\big)\cdot\alpha_i^1
\end{equation}

Satisfying (\ref{eq:ConceptDriftCondition sec:ConceptDriftGeneration}) exactly is computationally intractable, thus we stop iterating when the obtained error is close enough to the desired one. We say they are close enough when the absolute relative difference is less than 5\%, i.e. $|\frac{E_j-(1+\varepsilon)E(X_{i-1},C_{i-1})}{(1+\varepsilon)E(X_{i-1},C_{i-1})}|<0.05$.

\section{Proofs} \label{app:pro}

\begin{proof}[Proof of \textbf{Theorem 1}]
$\textbf{x}_i^t$ is a r.v. distributed according to $p_t$ and with support in $\mathbb{R}^d$, for $i=1,...,N$ and $t\geq 0$, where $p_t=p'$ for $t\geq T$ and $p_t=p$ for $T> t \geq 0$. Let us define the random variables $V_i^t=1/N\cdot(\mathbf{1}_{t< T}/T-\rho^t\cdot (1-\rho))\|\textbf{x}_i^t-\textbf{c}\|^2$, for $t\geq 0$. Let $\bar{V}= \sum_{i=1}^N \sum_{t\geq 0} V^t$, then $\bar{V}= E_{SKM} - E_\rho$, and $\mathbb{E}[\bar{V}]= \rho^T \cdot \varepsilon \cdot E/ N$.

The range of the support of $V_i^t$ is $r^t= u/N\cdot(\mathbf{1}_{t< T}/T - \rho^{t}\cdot (1-\rho))$, for $i=1,...,N$ and $t\geq 0$. Thus we have that
\begin{equation}
\sum_{t\geq 0}\sum_{i=1}^N (r^t)^2= \frac{u^2}{N}\cdot \left( \frac{2\rho^{T}-1}{T} + \frac{1-\rho}{1+\rho}\right)
\end{equation}

For $\gamma>0$, by Hoeffding's inequality, we have that $Pr[|\bar{V} - \mathbb{E}[\bar{V}|]\geq \gamma] \leq \delta$, where
\begin{align}
Pr[\rho^{T}\cdot \varepsilon \cdot E -\gamma  <E_{SKM} - E_\rho < \rho^{T} \cdot \varepsilon \cdot E + \gamma]> 1 - \delta,\nonumber
\end{align}

Therefore, with at least probability $1-\delta$, we have that $E_{SKM}-E_\rho \in (\rho^{T}\cdot \varepsilon \cdot E -\gamma,\rho^{T}\cdot \varepsilon \cdot E + \gamma)$, where
\begin{equation}
\gamma= u \cdot \sqrt{\frac{(2\cdot \rho^{T} - 1)/T + (1 - \rho)/(1 + \rho)}{2\cdot N}\cdot \ln \frac{2}{\delta}},
\end{equation}
which concludes the proof.
\end{proof}

\begin{proof}[Proof of \textbf{Theorem 2}]
 First we show that:
 \begin{align}
E_\rho(\mathcal{X},C)&= \frac{1}{M_{\mathcal{X}}}\cdot \sum_{t\geq0}^{} \rho^t \cdot \sum_{x \in B^t} \|\textbf{x}-\textbf{c}_x\|^2\nonumber\\
&= \frac{1}{M_{\mathcal{X}}} \cdot \sum_{t\geq1}^{} \rho^t \cdot \sum_{x \in B^{t}} \|\textbf{x}-\textbf{c}_x\|^2 +\nonumber\\
&+\frac{1}{M_{\mathcal{X}}}\cdot\sum_{x \in B^{0}} \|\textbf{x}-\textbf{c}_x\|^2 \nonumber\\ 
&= \frac{M_{\mathcal{X}\setminus \{B^0\}}}{M_{\mathcal{X}}}\cdot E_{1}^\rho(C) + \frac{|B^0|}{M_{\mathcal{X}}} \cdot E^{0}(C) 
\end{align}
 Then observe that
\begin{align}\label{NotFullDevelopedBounda}
E_0^\rho(C)&= \frac{M_{\mathcal{X}\setminus \{B^0\}}}{M_{\mathcal{X}}} \cdot E_{1}^\rho(C) + \frac{|B^0|}{M_{\mathcal{X}}} \cdot E^{0}(C) = \nonumber\\
&= \frac{1}{M_{\mathcal{X}}} \cdot \sum_{k=1}^{K}\sum_{t\geq1}^{} \rho^t \cdot \sum_{x \in B^{t}\cap P^*_k} \|\textbf{x}-\textbf{c}_x\|^2 +\nonumber\\
&+\frac{1}{M_{\mathcal{X}}} \cdot \sum_{k=1}^{K}\sum_{x \in B^0\cap P^0_k} \|\textbf{x}-\textbf{c}_x\|^2 \leq \nonumber\\
&\leq\frac{1}{M_{\mathcal{X}}} \cdot \sum_{k=1}^{K}\sum_{t\geq1}^{} \rho^t \cdot \sum_{x \in B^{t}\cap P^*_k} \|\textbf{x}-\textbf{c}_{k'}\|^2 +\nonumber\\
&+\frac{1}{M_{\mathcal{X}}} \cdot \sum_{k=1}^{K}\sum_{x \in B^0\cap P^0_k} \|\textbf{x}-\textbf{c}_{k''}\|^2\ 
\end{align}

Note that the last inequality holds as a consequence of the definition of $\textbf{c}_x$, while equality would hold if there were no reassignments. We compute $\textbf{c}_{k'}$ and $\textbf{c}_{k''}$ as the closest centroids from $C$ to the previous centroids $\textbf{c}^*_{k}$ and the new centroids $\textbf{c}^0_{k}$ respectively. In order to obtain the desired form of the upper-bound of $E_0^\rho(C)$, we shall recall how the centroids $\textbf{c}^*_{k}$ and $\textbf{c}^0_{k}$ are computed. With our notation, $\overline{B\cap P}$ is the mean value of the points in the set $B\cap P$.

\begin{align}\label{PrevNewCentersa}
\textbf{c}_{k}^{*}&=\frac{\sum_{t\geq1}^{} \rho^t \cdot |B^{t}\cap P_{k}^{*}|\cdot \overline{B^{t}\cap P_{k}^{*}}}{\sum_{t\geq1}^{} \rho^t \cdot |B^{t}\cap P_{k}^{*}|}\\ 
\textbf{c}_{k}^{0}&= \overline{B^{0}\cap P_{k}^{0}}\nonumber
\end{align}

On the other hand, using the identity\footnote{Knowing that this equation is true, it is quite straight forward to prove that it is also true for a weighted version.} $\sum_{x\in X}\|\textbf{x}-\textbf{c}\|^2=\sum_{x\in X}\|\textbf{x}-\overline{X}\|^2+|X|\|\overline{X}-\textbf{c}\|^2$, we obtain:
\begin{align}
&\frac{1}{M_{\mathcal{X}}} \cdot \sum_{k=1}^{K}\sum_{t\geq1}^{} \rho^t \cdot \sum_{x \in B^{t}\cap P^*_k} \|\textbf{x}-\textbf{c}_{k'}\|^2=\nonumber\\
&=\frac{1}{M_{\mathcal{X}}} \cdot \sum_{k=1}^{K}\sum_{t\geq1}^{} \rho^t \cdot \big(\sum_{x \in B^{t}\cap P^*_k} \|\textbf{x}-\overline{B^{t}\cap P^*_k}\|^2+\nonumber\\
&+|B^{t}\cap P^*_k|\cdot \|\overline{B^{t}\cap P^*_k}-\textbf{c}_{k'}\|^2\big)\nonumber\\
&=\frac{1}{M_{\mathcal{X}}} \cdot \sum_{k=1}^{K}\sum_{t\geq1}^{} \rho^t \cdot |B^{t}\cap P^*_k|\|\overline{B^{t}\cap P^*_k}-\textbf{c}_{k'}\|^2+const\nonumber
\end{align}

Note that the first term is independent of $\textbf{c}_{k'}$, so it is constant. Now we can develop the remaining term:
\vspace{-1.5mm}
\begin{align}
&\frac{1}{M_{\mathcal{X}}} \cdot \sum_{k=1}^{K}\sum_{t\geq1}^{} \rho^t \cdot |B^{t}\cap P^*_k|\cdot \|\overline{B^{t}\cap P^*_k}-\textbf{c}_{k'}\|^2+const=\nonumber\\
&=\frac{1}{M_{\mathcal{X}}} \cdot \sum_{k=1}^{K}\sum_{t\geq1}^{} \rho^t \cdot |B^{t}\cap P^*_k|\cdot (\|\overline{B^{t}\cap P^*_k}\|^2-\nonumber\\
&-2\cdot \overline{B^{t}\cap P^*_k}\cdot\textbf{c}_{k'}+\|\textbf{c}_{k'}\|^2)+const
\end{align}

from (\ref{PrevNewCentersa}) $\ \sum_{t\geq1}^{} \rho^t \cdot |B^{t}\cap P_{k}^{0}|\cdot \overline{B^{t}\cap P_{k}^{*}}=\sum_{t\geq1}^{} \rho^t \cdot |B^{t}\cap P_{k}^{*}|\cdot\textbf{c}_{k}^{*}$ holds, hence:
\vspace{-1.5mm}
\begin{align}
&\frac{1}{M_{\mathcal{X}}} \cdot \sum_{k=1}^{K}\sum_{t\geq1}^{} \rho^t \cdot |B^{t}\cap P^*_k|\cdot (\|\textbf{c}_{k'}\|^2-2\cdot \textbf{c}^*_k\cdot\textbf{c}_{k'}+\nonumber\\
&+\|\textbf{c}^*_{k}\|^2-\|\textbf{c}^*_{k}\|^2+\|\overline{B^{t}\cap P^*_k}\|^2)+const=\nonumber\\
&=\frac{1}{M_{\mathcal{X}}} \cdot \sum_{k=1}^{K}\sum_{t\geq1}^{} \rho^t \cdot |B^{t}\cap P^*_k|\cdot (\|\textbf{c}_{k'}-\textbf{c}^*_{k}\|^2+\nonumber\\
&+\|\overline{B^{t}\cap P^*_k}\|^2-\|\textbf{c}^*_{k}\|^2)+const=\nonumber\\
&=\frac{1}{M_{\mathcal{X}}} \cdot \sum_{k=1}^{K}\sum_{t\geq1}^{} \rho^t \cdot |B^{t}\cap P^*_k|\cdot\|\textbf{c}_{k'}-\textbf{c}^*_{k}\|^2+const\nonumber
\end{align}

In the last step we use the fact that $\|\overline{B^{t}\cap P^*_k}\|^2-\|\textbf{c}^*_{k}\|^2$ is independent of the centroids $\textbf{c}_{k'}$, therefore it is constant. The same reasoning can be applied for the second term in (\ref{NotFullDevelopedBounda}), but considering $\textbf{c}_{k}^{0}= \overline{B^{0}\cap P_{k}^{0}}$. Finally, combining both results we obtain our desired formula:
\vspace{-1.5mm}
\begin{align}
E_0^\rho(C)&\leq\frac{1}{M_{\mathcal{X}}}\sum_{k=1}^{K}\big[(\sum_{t\geq1}^{} \rho^t \cdot |B^{t}\cap P_{k}^{*}|)\cdot \|\textbf{c}_{k'}-\textbf{c}^*_k\|^2+\nonumber\\
&+|B^{0}\cap P_{k}^{0}|\|\textbf{c}_{k''}-\textbf{c}^0_k\|^2\big]+const;
\end{align}
 which concludes the proof.
\end{proof}

Here we have proved that the surrogate error is upper-bounded by a weighted $K$-means error over the set of previous and new centroids $\{\textbf{c}^*_{k},\textbf{c}^0_{k}\}$, plus a constant. The key component in this proof is the identity $\sum_{x\in X}\|\textbf{x}-\textbf{c}\|^2=\sum_{x\in X}\|\textbf{x}-\overline{X}\|^2+|X|\|\overline{X}-\textbf{c}\|^2 $, since we are able to remove the dependence of the error over the data points $\textbf{x}$ in this manner. If there are no reassignments, the inequality (\ref{NotFullDevelopedBounda}) turns into equality, and the error function does not depend on the data points explicitly. However, this is a strong assumption we make, which may not be true generally, because some data points may lie near cluster boundaries. 

\begin{proof}[Proof of \textbf{Theorem 3}]

First we take derivative of $f^\rho(\mathcal{X},C)$ with respect to each centroid $\textbf{c}_k$, and we equal it to 0:

\begin{align}
 \nabla_{\textbf{c}_k}f^{\rho}(C) =\frac{2}{M_{\mathcal{X}}}\big[w^*_k\cdot (\textbf{c}_{k}-\textbf{c}^*_k)+w^0_{\sig}\cdot(\textbf{c}_{k}-\textbf{c}^0_{\sig})\big]=0\nonumber
\end{align}

It is easy to prove from this equation that the set of centroids which minimizes (\ref{SimpleUpperBound}) has the form:

\begin{align}\label{WeightsEquationa}
\textbf{c}_k = \frac{1}{w^*_k+w^0_{\sig}}(w^*_k\cdot\textbf{c}^*_k+w^0_{\sig}\cdot\textbf{c}^0_{\sig})
\end{align}

Moreover, it is guaranteed to be always a minimum, since the associated Hessian matrix is positive definite.

\begin{align}
\nabla_{\textbf{c}_k;\textbf{c}_{m}}^2f^{\rho}(C)=\frac{2}{M_{\mathcal{X}}}\cdot(w^*_k+w^0_{\sig})\cdot\delta_{k,m}\geq 0\nonumber
\end{align}

Hence, the centroids $\textbf{c}_k$ of the form \ref{WeightsEquationa} minimize $f^\rho(\mathcal{X},C)$, which concludes the proof.
\end{proof}

\bibliographystyle{ieeetr}
\bibliography{Main.bib}

\end{document}